\documentclass[10pt,twocolumn,letterpaper]{article}

\usepackage{iccv}
\usepackage{times}
\usepackage{epsfig}
\usepackage{graphicx}
\usepackage{amsmath}
\usepackage{amssymb}

\usepackage{booktabs}
\usepackage{algorithm} 
\usepackage{algpseudocode}
\usepackage{mathrsfs}
\usepackage{multirow}
\usepackage{color}
\usepackage{colortbl}
\usepackage{arydshln}
\usepackage{array}
\usepackage{bm}
\usepackage{rotating}
\usepackage{adjustbox}
\usepackage{color}

\usepackage[pagebackref=true,breaklinks=true,letterpaper=true,colorlinks,bookmarks=false]{hyperref}

\iccvfinalcopy 

\def\iccvPaperID{2985} 

\ificcvfinal\fi

\begin{document}

\title{Efficient Attention Network: Accelerate Attention by Searching Where to Plug}

\author{Zhongzhan Huang\textsuperscript{\rm 1}\thanks{Equal contribution},
Senwei Liang\textsuperscript{\rm 2}\footnotemark[1],
Mingfu Liang\textsuperscript{\rm 3}\thanks{Equal contribution}, Wei He\textsuperscript{\rm 4}\footnotemark[2],  Haizhao Yang \textsuperscript{\rm 2}\thanks{Corresponding author} \\
\textsuperscript{\rm 1}Tsinghua University,
\textsuperscript{\rm 2}Purdue University,\\
\textsuperscript{\rm 3}Northwestern University,
\textsuperscript{\rm 4}Nanyang Technological University, \\
hzz\_dedekinds@foxmail.com,
liang339@purdue.edu,\\
mingfuliang2020@u.northwestern.edu, wei005@e.ntu.edu.sg\\
haizhao@purdue.edu}

\maketitle
\ificcvfinal\thispagestyle{empty}\fi

\begin{abstract}

Recently, many plug-and-play self-attention modules are proposed to enhance the model generalization by exploiting the internal information of deep convolutional neural networks~(CNNs). Previous works lay an emphasis on the design of attention modules for specific functionality. However, they ignore the importance of where to plug in the attention module since they connect the modules individually with each block of the entire CNN backbone for granted, leading to incremental computational cost and the number of parameters with the growth of network depth. Thus, we propose a method called Efficient Attention Network~(EAN) to improve the efficiency of the existing attention modules. In EAN, we share the attention module within the backbone and propose a novel reinforcement-learning-based method to search where to connect the shared attention module. Finally, we obtain the attention network with sparse connections between the backbone and modules, while~(1) accelerating inference~(2) reducing extra parameter increment, and~(3) maintaining accuracy. Extensive experiments on widely-used benchmarks and popular attention networks show the effectiveness of EAN. Furthermore, we empirically illustrate that our EAN has the capacity of transferring to other tasks and capturing informative features. The code is available at \url{https://github.com/gbup-group/EAN-efficient-attention-network}.
\end{abstract}

\section{Introduction}
\label{sec:1}
\begin{figure}[t]
    \centering
    \includegraphics[width=1\linewidth]{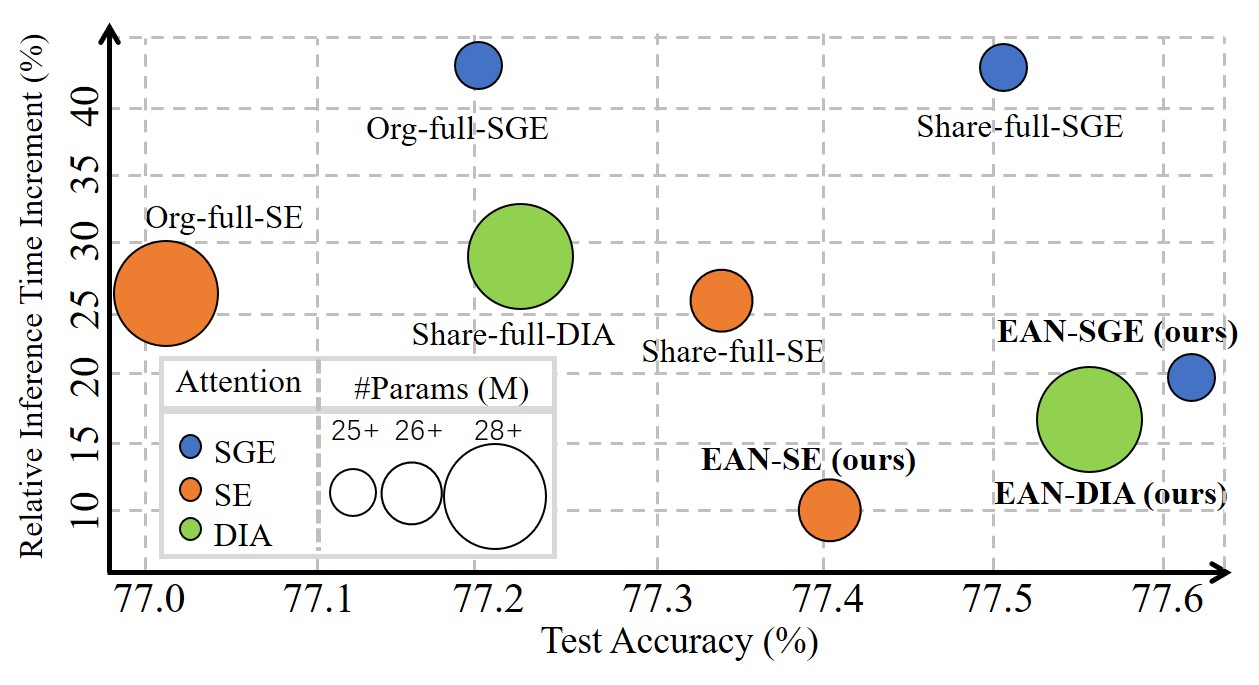}
    \caption{Comparison of relative inference time increment (see Eqn.~\ref{eqn:inference}), number of parameters, and test accuracy between various attention models on ImageNet 2012. We use different colors to distinguish the type of attention models, and the larger circle size means the larger number of parameters. Our models~(EAN) achieve the smaller relative inference time increment, parameters, and higher test accuracy than the attention models of same type.}
    \label{fig:qpt}
\end{figure}

Recently, many plug-and-play and straightforward self-attention modules that utilize the interior information of a network to enhance instance specificity~\cite{liang2020instance} are proposed to boost the generalization of deep convolutional neural networks~(CNNs)~\cite{hu2018squeeze,woo2018cbam,li2019spatial,huang2020dianet,cao2019gcnet,wang2018non}.
\begin{figure*}[ht]
\centering
\includegraphics[width=0.9\linewidth]{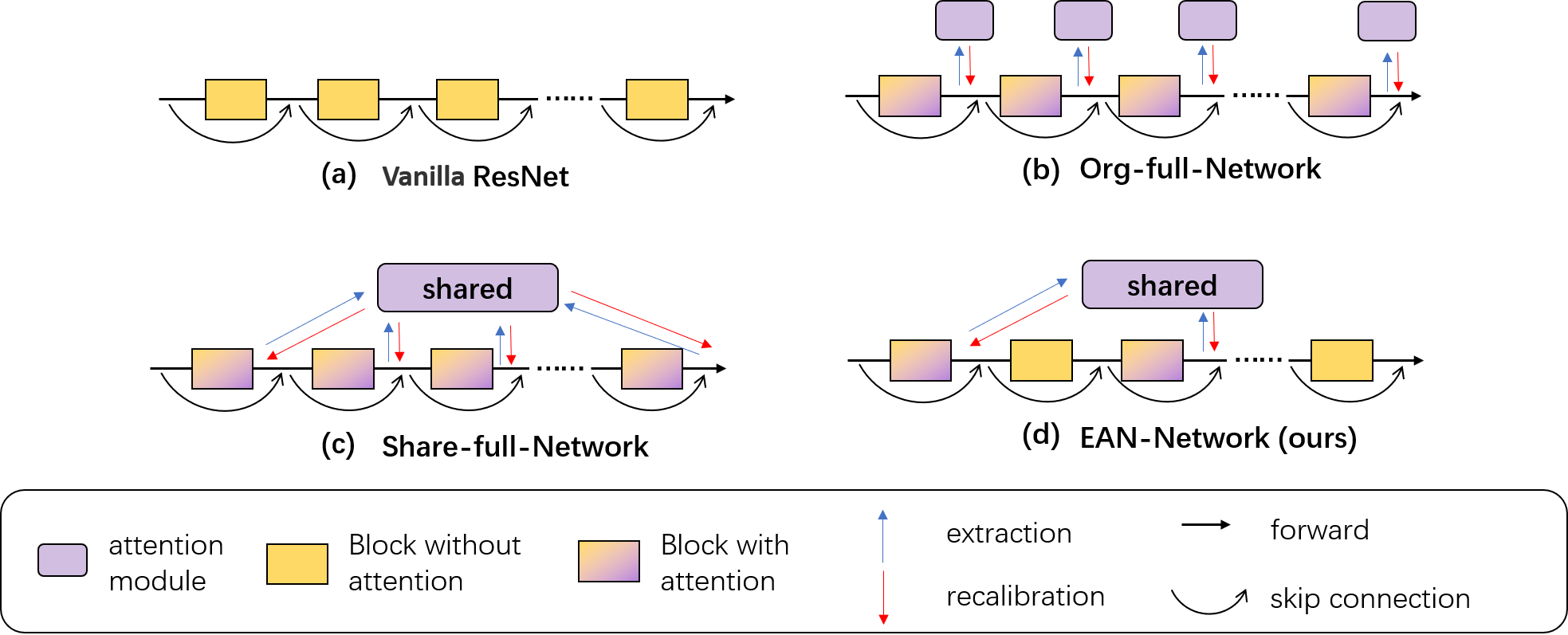}
\caption{Comparison of network structures between (a) ResNet, (b) Org-full attention network, (c) Share-full attention network, and (d) our EAN network. The detailed introduction of different networks is shown in Section~\ref{sec:prelim}.}
\label{fig:comparsion}
\end{figure*}
The self-attention module is usually plugged into every block of a residual network~(ResNet)~\cite{he2016deep}
~(see Fig.~\ref{fig:comparsion} (a) for the structure of a ResNet and Fig.~\ref{fig:comparsion} (b) for a network with attention modules). In general, the implementation of the attention module can be divided into three steps~\cite{huang2020dianet}: \textbf{(1)~Extraction}: the plug-in module extracts internal features of a network by computing their statistics, like mean, variance, or higher-order moments~\cite{hu2018squeeze,lee2019srm}; \textbf{(2)~Processing}: the module leverages the extracted features to adaptively generate a mask that measures the importance of the feature maps via a fully connected layer~\cite{hu2018squeeze}, convolution layer~\cite{woo2018cbam}, or feature-wise linear transformation~\cite{liang2020instance,li2019spatial} etc.; \textbf{(3)~Recalibration}: the mask is used to calibrate the feature maps of the network by element-wise multiplication or addition~\cite{hu2018squeeze,cao2019gcnet}. The operations and trainable modules in the implementation of self-attention inevitably require extra computational cost and parameters, resulting in slow inference and cumbersome network~\cite{bianco2018benchmark}. This limits self-attention usage on applications that need a real-time response or small memory consumption, such as robotics, self-driving car, and mobile device.

Previous works mainly focus on improving the capacity of self-attention module~\cite{woo2018cbam,huang2020dianet} or reducing the parameters~\cite{liang2020instance,li2019spatial,huang2020dianet}. However, these methods do not balance the inference time and parameters$/$performance. For example, Huang et al.~\cite{huang2020dianet} propose sharing mechanism to share an attention module with the same set of parameters to different blocks in the same stages as Fig.~\ref{fig:comparsion} (c). Though the shared module significantly reduces the trainable parameters, the computational cost remains the same. Besides, the lightweight attention design, such as feature-wise linear transformation used in SGE~\cite{li2019spatial}, reduces the parameters of individual modules and achieves good performance, but it still increases inference time as in Fig.~\ref{fig:qpt}. Naturally, a question would be asked:

\vspace{0.22cm}
\textit{Can we modify the existing self-attention networks such that the modified networks can (1) achieve faster inference speed, (2) contain fewer parameters, and (3) still maintain comparable accuracy at the same time?}
\vspace{0.22cm}

The major obstacle of accelerating the attention network is the connections between modules and the network. The previous implementation of attention modules follows conventional practice where the attention modules are individually plugged into every block throughout a CNN shown in Fig.~\ref{fig:comparsion} (b)~\cite{hu2018squeeze,woo2018cbam,li2019spatial,huang2020dianet,cao2019gcnet}, and hence the computational cost increases with the growing number of blocks. However, how many modules should be used and where to plug the module are rarely discussed.

To improve the efficiency of the attention modules in CNNs, in this paper, a simple idea is proposed to reduce the number of interactions between blocks and attention modules instead of plugging the attention modules into each block. Meanwhile, to reduce the parameter cost, we adopt the sharing mechanism~\cite{huang2020dianet}, and achieve the outcome in Fig.~\ref{fig:comparsion} (d). Comparing to Fig.~\ref{fig:comparsion} (b) and (c), our advantages are two-folded, \textcircled{1} smaller parameter increment \textcircled{2} less computational cost increment because of fewer connections between backbone and attention module.

However, to achieve satisfactory performance, the dense connections~\cite{huang2017densely} and the adequate number of trainable parameters~\cite{simonyan2014very} in networks are two critical factors in general. Thus, to balance efficiency and satisfactory performance, we propose a novel reinforcement-learning-based method to search for the optimal connection scheme. Compared with the popular architecture search methods~\cite{Vidnerov2020Multi,pham2018efficient,liu2018darts}, our proposed method can better achieve all the goals where we can obtain attention network with sparse connections between the backbone and modules while~(1) accelerating inference~(2) reducing extra parameter increment and~(3) maintaining accuracy. Our method is called Efficient Attention Network~(EAN), which shares the attention module within the backbone and searches where to connect the shared attention module.

\textbf{Our Contribution.}

    1. We propose an effective connection searching method to improve the efficiency of the various attention network while maintaining the original accuracy, reducing the extra parameters increment, and accelerating the inference.

    2. Our results show that disconnecting some modules from the backbone can be harmless or even improve the model performance. Such empirical finding differs from the intuition behind the conventional approach which applies attention modules to each block.
    
    3. Through our empirical experiments, we illustrate that the attention network searched by our method has the capacity of transferring to other tasks and capturing the informative features, which means our EAN networks have potential to apply to wide fields. 

\section{Related Works}
\textbf{Neural Architecture Search~(NAS).} Designing a satisfactory neural architecture automatically, also known as neural architecture search, is of significant interest for academics and industrial AI research. Such a problem may always be formulated as searching for the optimal combination of different network granularities.

The early NAS works require expensive computational cost for scratch-training a massive number of architecture candidates~\cite{zoph2016neural, zoph2018learning}. To alleviate the searching cost, the recent advances of one-shot approaches for NAS bring up the concept of supernet based on the weight-sharing heuristic. Supernet serves as the search space embodiment of the candidate architectures, and it is trained by optimizing different sub-networks from the sampling paths, \textit{e.g.}, SPOS~\cite{guo2020single}, GreedyNAS~\cite{you2020greedynas} and FairNAS~\cite{chu2019fairnas}.

The most conceptually related work~\cite{li2020neural} aims to propose a lightweight non-local~(LightNL) block~\cite{wang2018non} and searches for the optimal configuration to incorporate the non-local block into mobile neural networks. Although the inserted location of the LightNL is also considered in their NAS objective, the construction of the LightNL blocks is also jointly optimized in their objective. As both the inserted location and the construction of LightNL are integrated completely after the searching, it is hard to differentiate the net contribution of their proposed inserted location of LightNL blocks. However, in our work, we tailor to identify the importance of \textit{where to plug} in the attention module, and we do not concentrate on only one design of the existing attention modules, compared to Li et al.~\cite{li2020neural} that only specializes on non-local block~\cite{wang2018non}. To sum up, the difference in the research target and more general consideration differentiate our work with Li et al.~\cite{li2020neural}.

\textbf{Self-Attention Mechanism.} The self-attention mechanism is widely used in CNNs for computer vision~\cite{hu2018squeeze,wang2018non,huang2020dianet,cao2019gcnet,li2019spatial,liang2020instance}. The self-attention module is modularized as a network component and inserted into different layers of the network to emphasize informative features and their importance according to the internal information. 

Many works focus on the design of the attention module for specific functionality. Squeeze-and-Excitation~(SE) module~\cite{hu2018squeeze} leverages global average pooling to extract the channel-wise statistics and learns the non-mutually-exclusive relationship between channels. Spatial Group-wise Enhance~(SGE) module~\cite{li2019spatial} learns to recalibrate features by saliency factors learned from different groups of the feature maps. Dense-Implicit-Attention~(DIA) module~\cite{huang2020dianet} captures the layer-wise feature interrelation with a recurrent neural network~(RNN).

\section{Preliminaries}         
\label{sec:prelim}
In this section, we briefly review ResNet~\cite{he2016deep}. Then, we formulate two types of attention networks: Org-full network (Fig.~\ref{fig:comparsion} (b)), and Share-full network (Fig.~\ref{fig:comparsion} (c)). The structure of ResNet is shown in Fig.~\ref{fig:comparsion}~(a). In general, the ResNet architecture has several stages, and each stage, whose feature maps have the same size, is a collection of consecutive blocks. Suppose a ResNet has $m$ blocks. Let $x_\ell$ be the input of the $\ell^\text{th}$ block and $f_\ell(\cdot)$ be the residual mapping, and then the output $x_{\ell+1}$ of the $\ell^\text{th}$ block is defined as 
\begin{align}
    x_{\ell+1} = x_\ell + f_\ell(x_\ell).
\end{align}

\subsection{Org-full Attention Network} 
We describe an attention network as an Org-full network (Fig.~\ref{fig:comparsion} (b)) if the attention module is individually defined for each block. Note that the term ``full'' refers to a scenario that all blocks in a network connect to the attention modules, while ``Org'' is short for ``Original''. Many popular attention modules adopt this way to connect the ResNet backbone~\cite{hu2018squeeze,li2019spatial,woo2018cbam}. We denote the attention module in the $\ell^\text{th}$ block as $M(\cdot; W_\ell)$, where $W_\ell$ are the parameters. Then the attention will be formulated as $M(f_\ell(x_\ell);W_\ell)$ which consists of the extraction and processing operations introduced in Section~\ref{sec:1}. In the recalibration step, the attention is applied to the residual output $f_\ell(x_\ell)$, \textit{i.e.},
\begin{align}
    x_{\ell+1} = x_\ell + M(f_\ell(x_\ell);W_\ell)\odot f_\ell(x_\ell),
    \label{eqn:org-full}
\end{align}
where $\ell=1,\cdots, m$ and $\odot$ is the element-wise multiplication. Eqn.~\ref{eqn:org-full} indicates that the computational cost and number of parameters grow with the increasing number of blocks $m$. 

\subsection{Share-full Attention Network}
We denote an attention network as a Share-full network (Fig.~\ref{fig:comparsion} (c)) if the blocks within one stage are connected to the same attention module defined for the stage. 

The idea of Share-full network is first proposed in Huang et al.~\cite{huang2020dianet}. We denote attention module defined in the stage $k$ as $M(\cdot;W_{k})$. If the $\ell^\text{th}$ block belongs to the $k_\ell$ stage, then the attention is modeled as $M(f_\ell(x_\ell);W_{k_\ell})$. The building block becomes 
\begin{align}
    x_{\ell+1} = x_\ell + M(f_\ell(x_\ell);W_{k_\ell})\odot f_\ell(x_\ell),
    \label{eqn:share-full}
\end{align}
where $\ell=1,\cdots, m$. 

Distinct from the Org-full attention network, the number of extra parameters of the Share-full network depends on the number of stages, instead of the number of blocks $m$. Typically, a ResNet has 3$\sim$4 stages but has tens of blocks, which indicates a Share-full network can significantly reduce the extra parameters.

\section{Proposed Method}
In this section, we systematically introduce the proposed Efficient Attention Network~(EAN) method, which consists of two parts: First, we pre-train a supernet as the search space, and the supernet has the same network structure as a Share-full network. Second, we use a policy-gradient-based method to search for an optimal connection scheme from the supernet. The basic workflow of our method is shown in Alg.~\ref{alg:ean}. 

\begin{algorithm}[t]  
    \caption{Searching optimal connection scheme}\label{alg:ean}   
    \textbf{Input:} Training set $D_\text{train}$; validation set $D_\text{val}$; a Share-full network $\Omega(\mathbf{x}|\mathbf{1})$; learning rate $\eta$; pre-training step $K$; searching step $T$; time step $h$ to apply PPO.
    
    \textbf{Output:} The trained controller $\chi_\theta(x_0)$.
        
    \begin{algorithmic}[1]

    \State\algorithmiccomment{Pre-train the supernet}

    \For{$t$ from 1 to $K$} 

        \State $\mathbf{a} \sim [Bernoulli(0.5)]^m$
        \State train $\Omega(\mathbf{x}|\mathbf{a})$ with $D_\text{train}$
    \EndFor  
    
    \State\algorithmiccomment{Policy-gradient-based search}       
       \For{$t$ from 1 to $T$} 

        \State $\mathbf{p}_\theta \gets \chi_\theta(x_0)$
        \State $\mathbf{a} \sim \mathbf{p}_\theta$
        \State $g_\text{spa} \gets$ Eqn.~\ref{eqn:sparse}
        \State $g_\text{val} \gets \Omega(D_\text{val}|\mathbf{a})$, $g_\text{rnd}$ $\gets$ $\left\Vert\sigma_1(\mathbf{a})-\sigma_2(\mathbf{a};\phi)\right\Vert_2^2$
        \State  calculate the reward $G(\mathbf{a})$ by Eqn.~\ref{eqn:rnd_reward}
        \State update $\theta$ by Eqn.~\ref{eqn:policy_gradient}
        \State update $\phi$ by minimizing $\left\Vert\sigma_1(\mathbf{a})-\sigma_2(\mathbf{a};\phi)\right\Vert_2^2$
        \State put $\left(\mathbf{p}_\theta,\mathbf{a},G(\mathbf{a})\right)$ into replay buffer
       
          \State\algorithmiccomment{Update $\theta$ from buffer}   
        \If{t $\geq$ h}
            \State sample $\left(\mathbf{p}_\theta,\mathbf{a},G(\mathbf{a})\right)$ from replay buffer
            \State update $\theta$ by Eqn.~\ref{eqn:ppo}

        \EndIf
    
    \EndFor
    \State\Return $\chi_\theta(x_0)$
    \end{algorithmic} 

\end{algorithm}  

\subsection{Problem Description}
We consider a supernet $\Omega(\mathbf{x}|\mathbf{a})$ with $m$ blocks and input $\mathbf{x}$, and $\Omega(\mathbf{x}|\mathbf{a})$ has the same network structure as a Share-full network. A sequence $\mathbf{a} = (a_1,a_2,\cdots,a_m)$ denotes an attention connection scheme, where $a_i =1$ when the $i^\text{th}$ block is connected to the shared attention module, otherwise it is equal to 0. A sub-network specified by a scheme $\mathbf{a}$ can be formulated as follows:
\begin{align}
\begin{split}
    x_{\ell+1} = x_\ell + \Big( a_\ell\cdot M(f_\ell(x_\ell);W_{k_\ell})&+(1-a_\ell)\mathbf{1} \Big) \odot f_\ell(x_\ell), 
\end{split}
\end{align}
where $\mathbf{1}$ denotes an all-one vector and $\ell$ is from $1$ to $m$. In particular, $\Omega(\mathbf{x}|\mathbf{a})$ becomes a Share-full network if $\mathbf{a}$ is all-one vector, or a vanilla ResNet while $\mathbf{a}$ is a zero vector. Our goals are: (1) to find a connection scheme $\mathbf{a}$, which is sparse enough for less computation cost, from $2^m$ possibilities; (2) to ensure that the network $\Omega(\mathbf{x}|\mathbf{a})$ possesses good generalization.

\subsection{Pre-training the Supernet}
\label{sec:pretrain}

To determine the optimal architecture from the pool of candidates, it is costly to evaluate all their individual performance after training. In many related works on NAS, candidates' validation accuracy from a supernet serve as a satisfactory performance proxy~\cite{guo2020single,you2020greedynas,chu2019fairnas}. Similarly, to obtain the optimal connection scheme for the attention module, we propose to train the supernet as the search space following the idea of co-adaption~\cite{liang2018drop}. We consider the validation performance of the sampled sub-networks as the proxy for their stand-alone\footnote{Train the sub-networks from scratch} performance.

Specifically, given a dataset, we split all training samples into the training set $D_\text{train}$ and the validation set $D_\text{val}$. To train the supernet, we activate or deactivate the attention module in each block of it randomly during optimization. 

We first initialize a supernet $\Omega(\mathbf{x}|\mathbf{a}^{(0)})$, where $\mathbf{a}^{(0)} = (1,\cdots,1)$. At iteration $t$, we randomly draw a connection scheme $\mathbf{a}^{(t)}=(a^t_1, \cdots, a^t_m)$, where $a^t_i$ is sampled from a Bernoulli distribution $B(0.5)$. Since every attention module connects or disconnects to the block, without any prior knowledge, it is reasonable to choose probability $0.5$ for randomly drawing a connection scheme. Then, we train sub-network $\Omega(\mathbf{x}|\mathbf{a}^{(t)})$ with the scheme $\mathbf{a}^{(t)}$ from $\Omega(\mathbf{x}|\mathbf{a}^{(0)})$ on $D_\text{train}$ via weight-sharing. 


\begin{table*}[ht]
  \centering
  \begin{adjustbox}{width=\textwidth,center}
    \begin{tabular}{cllccclccclccc}
    \toprule
    \multirow{2}[4]{*}{Dataset} & \multicolumn{1}{c}{\multirow{2}[4]{*}{Model}} &       & \multicolumn{3}{c}{Test Accuracy (\%)} &       & \multicolumn{3}{c}{Parameters (M)} &       & \multicolumn{3}{c}{Relative Inference Time Increment (\%)} \\
\cmidrule{4-6}\cmidrule{8-10}\cmidrule{12-14}          &       &       & Org-full & Share-full & EAN   &       & Org-full & Share-full & EAN   &       & Org-full & Share-full & EAN  \\
    \midrule
    \multirow{4}{*}{\rotatebox{90}{CIFAR100}} & Org   &       & 74.29  & -     & -     &       & 1.727  & -     & -     &       & 0.00  & -     & - \\
          & SE~\cite{hu2018squeeze} &       & 75.80  & 76.09  & \textbf{76.93} &       & 1.929  & 1.739  & \textbf{1.739} &       & 54.46  & 52.09  & \textbf{23.52}~({\color{red}{$\downarrow$ 28.57}}) \\
          & SGE~\cite{li2019spatial} &       & 75.75  & 76.17  & \textbf{76.36} &       & 1.728  & 1.727  & \textbf{1.727} &       & 93.60  & 93.41  & \textbf{50.49}~({\color{red}{$\downarrow$ 42.92}}) \\
          & DIA~\cite{huang2020dianet} &       & -     & \textbf{77.26} & 77.12  &       & -     & 1.946  & \textbf{1.946} &       & -     & 121.11  & \textbf{65.46}~({\color{red}{$\downarrow$ 55.65}}) \\
    \midrule
    \multirow{4}{*}{\rotatebox{90}{ImageNet}} & Org   &       & 76.01  & -     & -     &       & 25.584  & -     & -     &       & 0.00  & -     & - \\
          & SE~\cite{hu2018squeeze}    &       & 77.01  & 77.35  & \textbf{77.40} &       & 28.115  & 26.284  & \textbf{26.284} &       & 25.94  & 25.92  & \textbf{10.35}~({\color{red}{$\downarrow$ 15.57}}) \\
          & SGE~\cite{li2019spatial} &       & 77.20  & 77.51  & \textbf{77.62} &       & 25.586  & 25.584  & \textbf{25.584} &       & 40.60  & 40.50  & \textbf{19.66}~({\color{red}{$\downarrow$ 20.84}}) \\
          & DIA~\cite{huang2020dianet}   &       & -     & 77.24  & \textbf{77.56} &       & -     & 28.385  & \textbf{28.385} &       & -     & 27.26  & \textbf{16.58}~({\color{red}{$\downarrow$ 10.68}}) \\
    \bottomrule
    \end{tabular}%
    \end{adjustbox}
    \caption{Comparison of relative inference time increment (see Eqn.~\ref{eqn:inference}), number of parameters, and test accuracy between various attention models on CIFAR100 and ImageNet 2012. ``Org'' stands for ResNet164 backbone in CIFAR100 and ResNet50 backbone in ImageNet. EAN networks have faster inference speed among the networks with the same type of attention module compared with the same type Share-full attention network.}
  \label{tab:all_results}%
\end{table*}%


\begin{figure}[t]
\includegraphics[width=1.0\linewidth]{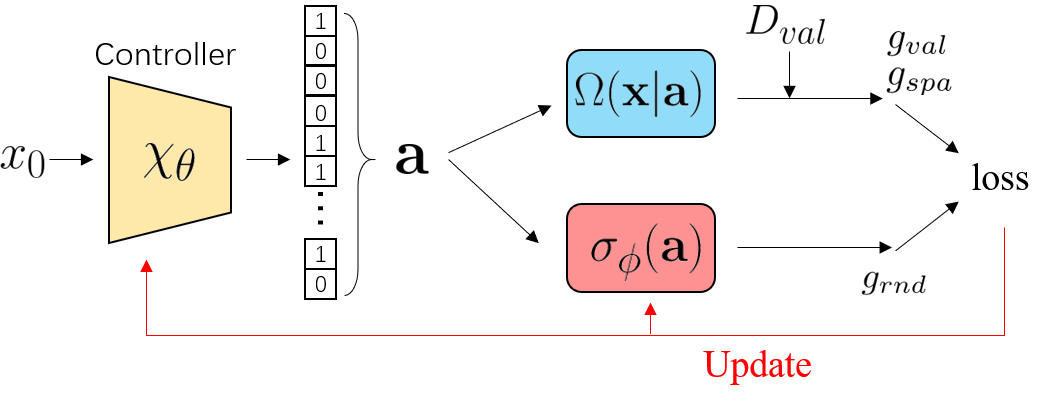}
\caption{The illustration of our policy-gradient-based method to search an optimal scheme.}
\label{fig:arch}
\end{figure}

\subsection{Training Controller with Policy Gradient}
In this part, we introduce the step to search for the optimal connection scheme. Concretely, we use a controller to generate connection schemes and update the controller by policy gradient, as illustrated in Fig.~\ref{fig:arch}. 

We use a fully connected network as controller $\chi_\theta(x_0)$ to produce the connection schemes, where $\theta$ are the learnable parameters, and $x_0$ is a constant vector $\mathbf{0}$. 
The output of $\chi_\theta(x_0)$ is $\mathbf{p_\theta}$, where $\mathbf{p_\theta} = (p_\theta^1,p_\theta^2,...,p_\theta^m)$ and $p_\theta^i$ represents the probability of connecting the attention to the $i^\text{th}$ block. A realization of $\mathbf{a}$ is sampled from the controller output, \textit{i.e.}, $\mathbf{a} \sim \mathbf{p}_\theta$. The probability associated with the scheme $\mathbf{a}$ is $\mathbf{\hat{p}_\theta} = (\hat{p}_\theta^1,\hat{p}_\theta^2,...,\hat{p}_\theta^m)$, where $\hat{p}_\theta^i = (1-a_i)(1-p_\theta^i)+a_ip_\theta^i$. 

We denote $G(\mathbf{a})$ as a reward for $\mathbf{a}$. The parameter set $\theta$ can be updated via policy gradient with learning rate $\eta$, \textit{i.e.}, 
\begin{align}
\begin{split}
    R_\theta &= G(\mathbf{a})\cdot\sum_{i=1}^m\log \hat{p}_\theta^i,\\
     \theta &= \theta + \eta\cdot \nabla {R}_\theta.
\end{split}
    \label{eqn:policy_gradient}
\end{align}

In this way, the controller tends to output the probability that results in a large reward $G$. Therefore, designing a reasonable $G$ can help us search for a good structure. 

\textbf{Sparsity Reward.} One of our goals is to accelerate the inference of the attention network. To achieve, we complement a sparsity reward $g_\text{spa}$ to encourage the controller to generate the schemes with fewer connections between attention modules and backbone. We define $g_\text{spa}$ by 
\begin{align}
    g_\text{spa} =  1 - \frac{\left\Vert \mathbf{a}\right\Vert_0}{m},
    \label{eqn:sparse}
\end{align}where $\left\Vert\cdot\right\Vert_0$ is a zero norm that counts the number of non-zero entities, and $m$ is the number of blocks.

\textbf{Validation Reward.} The other goal is to find the schemes with which the networks can maintain the original accuracy. Hence, we use the validation accuracy of the sub-network $\Omega(\mathbf{x}|\mathbf{a})$ sampled from the supernet as a reward, which depicts the performance of its structure. The accuracy of $\Omega(\mathbf{x}|\mathbf{a})$ on $D_\text{val}$ is denoted as $g_\text{val}$. In fact, it is popular to use validation accuracy of a candidate network as a reward signal in NAS~\cite{pham2018efficient,zoph2016neural,guo2020single,zoph2018learning,you2020greedynas}. Furthermore, it has been empirically proven that the validation performance of the sub-networks sampled from a supernet can be positively correlated to their stand-alone performance~\cite{bender2018understanding}. We evaluate the correlation between the validation accuracy of subnetworks sampled from a supernet and their stand-alone performance on CIFAR100 with ResNet and SE module over 42 samples and obtain the Pearson coefficient is 0.71.


\textbf{Curiosity Bonus.} 
To encourage the controller to explore more potentially useful connection schemes, we add the Random Network Distillation~(RND) curiosity bonus~\cite{burda2018exploration} in our reward. 
Two extra networks with input $\mathbf{a}$ are involved in the RND process, including a target network $\sigma_1(\cdot)$ and a predictor network $\sigma_2(\cdot;\phi)$, where $\phi$ is the parameter set. The parameters of $\sigma_1(\cdot)$ are randomly initialized and fixed after initialization, while $\sigma_2(\cdot;\phi)$ is trained with the connection schemes collected by the controller. 

The basic idea of RND is to minimize the difference between the outputs of these two networks, which is denoted by term $\sigma_\phi(\cdot) = \left\Vert\sigma_1(\cdot)-\sigma_2(\cdot;\phi)\right\Vert_2^2$, over the seen connection schemes.
If the controller generates a new scheme $\mathbf{a}$, $\sigma_\phi(\mathbf{a})$ is expected to be larger because the predictor $\sigma_2(\cdot;\phi)$ never trains on scheme $\mathbf{a}$.
Then, we denote the term $\left\Vert\sigma_1(\mathbf{a})-\sigma_2(\mathbf{a};\phi)\right\Vert_2^2$ as $g_\text{rnd}$, which is used as curiosity bonus to reward the controller for exploring a new scheme. Besides, in Fig.~\ref{fig:enas_vs_ean}, we empirically show that RND bonus mitigates the fast convergence of early training iterations, leading to exploration for more schemes.

To sum up, our reward $G(\mathbf{a})$ becomes
\begin{align}
    G(\mathbf{a}) = \lambda_1\cdot g_\text{spa} + \lambda_2 \cdot
    g_\text{val}+ \lambda_3 \cdot g_\text{rnd}, 
    \label{eqn:rnd_reward}
\end{align}
where $\lambda_1, \lambda_2, \lambda_3$ are the coefficient for each bonus. 

\textbf{Data Reuse.} 

To improve the utilization efficiency of sampled connection schemes and speed up the training of the controller, we incorporate Proximal Policy Optimization~(PPO)~\cite{schulman2017proximal} in our method. As shown in Alg.~\ref{alg:ean}, after the update of parameter $\theta$ and $\phi$, we put the tuple $(\mathbf{p}_\theta,~\mathbf{a},~G(\mathbf{a}))$ into a buffer. At the later step, we retrieve some used connection scheme and update $\theta$ as follows: 
\begin{align}
\begin{split}
    \kappa&=\mathbb{E}_{\mathbf{a}\sim\mathbf{p}_{\theta_{old}}}\left[G(\mathbf{a})\sum_{i=1}^m\frac{\hat{p}^i_\theta}{\hat{p}^i_{\theta_{old}}}\nabla_\theta\log \hat{p}_\theta^i\right],\\
    \theta&=\theta+\eta\cdot\kappa,
\end{split}
    \label{eqn:ppo}
\end{align}
 where $\eta$ is learning rate and the $\theta_{old}$ denotes the $\theta$ sampled from buffer. 

\section{Experiments}
\subsection{Datasets and Settings} 
 On CIFAR100 \cite{cifar} and ImageNet 2012~\cite{ILSVRC15} datasets, we conduct experiments on ResNet~\cite{he2016deep} backbone with different attention modules, including Squeeze-Excitation (SE) \cite{hu2018squeeze}, Spatial Group-wise Enhance (SGE) \cite{li2019spatial} and Dense-Implicit-Attention (DIA) \cite{huang2020dianet} modules. In our supplementary, we describe these modules as well as the training settings of controller and networks.
 
 Since the networks with attention modules have extra computational cost from the vanilla backbone inevitably, we formulate the relative inference time increment to represent the relative speed of different attention networks, \textit{i.e.},
 \begin{equation}
 \frac{I_t(w.~Attention) - I_t(wo.~Attention)}{I_t(wo.~Attention)} \times 100\%,
 \label{eqn:inference}
 \end{equation}
where $I_t(\cdot)$ denotes the inference time of the network and the notation $w/wo.~Attention$ represents the network with$/$without the attention module. The inference time is measured by forwarding the data of batch size 50 for 1000 times on a server with Intel(R) Xeon(R) Gold 5122 CPU @ 3.60GHz and 1 Tesla V100 GPU.

  \begin{figure}
    \centering
    \includegraphics[width=1\linewidth]{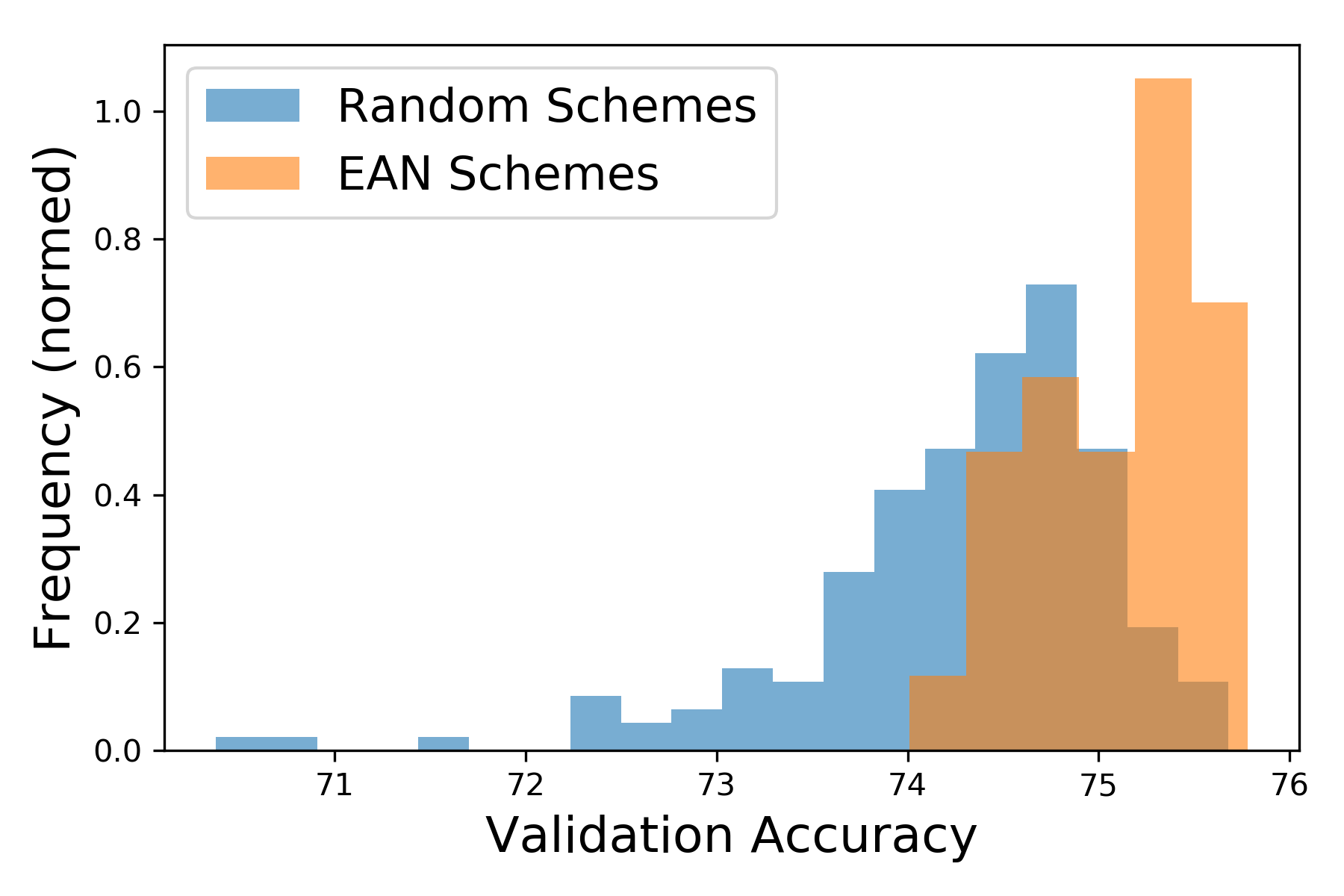}
    \caption{Comparison of the validation accuracy distribution between EAN and Random schemes for SE module. The validation accuracy is obtained by training from scratch the model on CIFAR100 with ResNet164 backbone.}
    \label{fig:rs_vs_ean}
\end{figure}
 
 \begin{figure}
    \centering
    \includegraphics[width=1\linewidth]{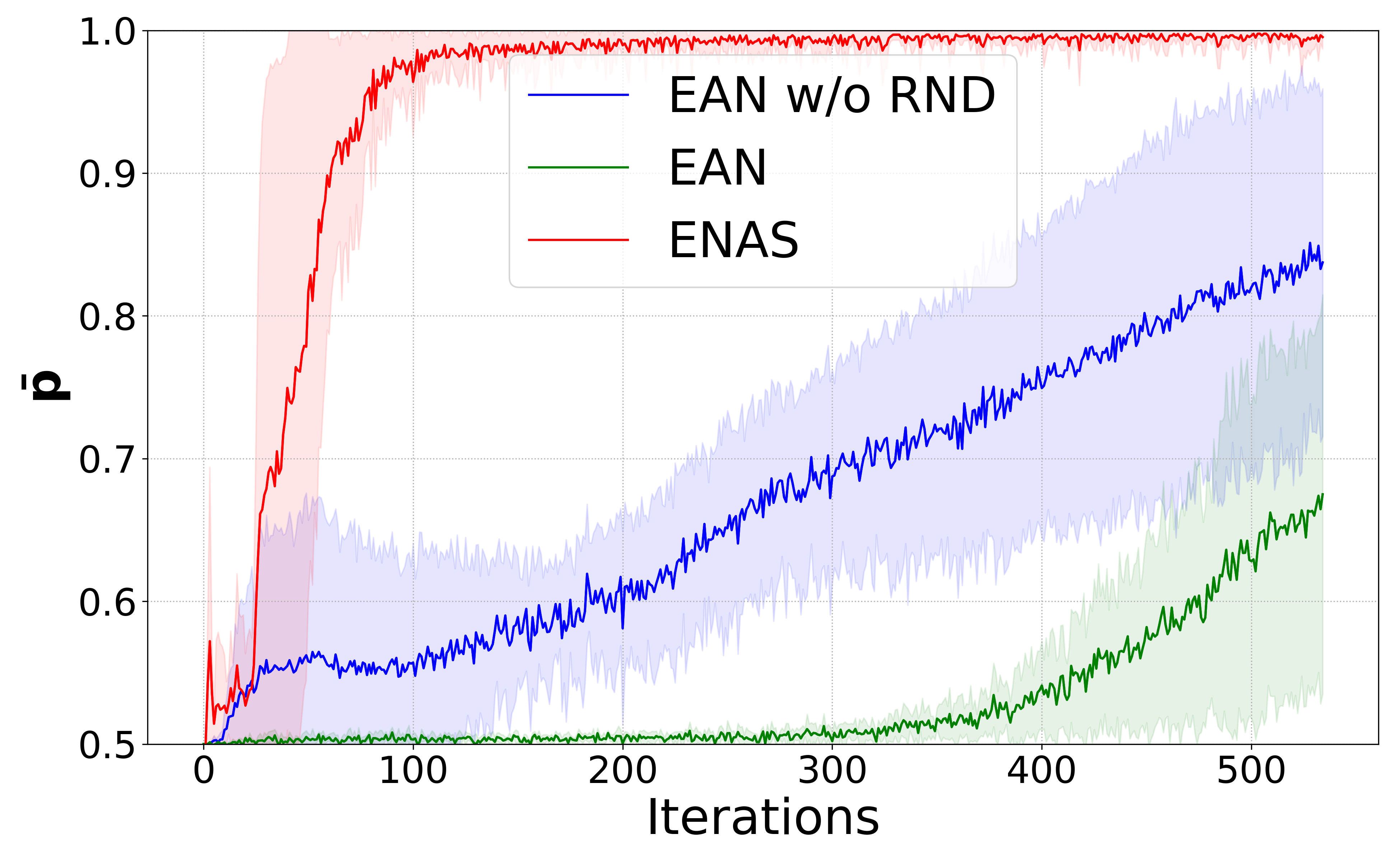}
    \caption{Comparison of the convergence speed between ENAS and EAN. The controller tends to generate a deterministic scheme when $\bar{\mathbf{p}}$ is close to 1.}
    \label{fig:enas_vs_ean}
\end{figure}
 
 \textbf{CIFAR100.} CIFAR100 consists of 50k training images and 10k test images of size 32 by 32. In our implementation, we choose 10k images from the training images as a validation set (100 images for each class, 100 classes in total), and the remainder images as a sub-training set. Regarding the experimental settings of ResNet164~\cite{he2016deep} backbone with different attention modules, the supernet is trained for 150 epochs, and the search step $T$ is set to be 1000. 
 
 \textbf{ImageNet 2012.} ImageNet 2012 comprises 1.28 million training images. We split 100k images (100 from each class and 1000 classes in total) as the validation set and the remainder as the sub-training set. The testing set includes 50k images. Besides, the random cropping of size 224 by 224 is used in ImageNet experiments. Regarding the experimental settings of ResNet50~\cite{he2016deep} backbone with different attention modules, the supernet is trained for 40 epochs, and the search step $T$ is set to be 300. 
 

 \begin{table*}[htbp]
  \centering
  \begin{adjustbox}{width=0.8\textwidth,center}
    \begin{tabular}{ccccc}
    \toprule
    Method & Stage1 & Stage2 & Stage3 & Test Accuracy (\%) \\
    \midrule
    ENAS (a) & 001001001001001001 & 001001001001001001 & 001001001001001001 & 75.80 \\
    ENAS (b) & 100100101100100100 & 101101100100101101 & 100100101101100100 & 75.11 \\
    ENAS (c) & 110110110110110110 & 110110110110110110 & 110110110110110110 & 76.08 \\
    \midrule
    EAN (a) & 001100100101110101 & 001100000111001111 & 101100000111110001 & \textbf{76.93} \\
    EAN (b) & 001100000001010111 & 011100001000010111 & 101000100110000000 & \textbf{76.71} \\
    \bottomrule
    \end{tabular}%
    \end{adjustbox}
  \caption{The connection schemes searched by ENAS~\cite{pham2018efficient} or EAN. The experiment is conducted on CIFAR100 with SE module and ResNet164 backbone.}
  \label{tab:ean_vs_enas}%
\end{table*}%
 
\begin{table*}[htbp]
  \centering
  \begin{adjustbox}{width=0.8\textwidth,center}
    \begin{tabular}{clcccccccl}
    \toprule
    \multirow{2}[4]{*}{Dataset} & \multicolumn{1}{c}{\multirow{2}[4]{*}{Model}} &       & \multicolumn{3}{c}{MAE/MSE} &       & \multicolumn{3}{c}{Relative Inference Time Increment (\%)} \\
\cmidrule{4-6}\cmidrule{8-10}          &       &       & Org-full & Share-full & EAN   &       & Org-full & Share-full & EAN \\
    \midrule
    \multirow{2}[2]{*}{SHHB} & SE~\cite{hu2018squeeze} &       & 9.5/15.93 & 8.9/14.6 & 8.6/14.7 &       & 19.19  & 19.19  & \textbf{6.16}~({\color{red}{$\downarrow$ 13.03}}) \\
          & DIA~\cite{huang2020dianet} &       & -     & 9.1/14.9 & 8.2/13.9 &       & -     & 16.93  & \textbf{8.71}~({\color{red}{$\downarrow$ 8.22}}) \\
    \midrule
    \multirow{3}[2]{*}{SHHA} & SGE~\cite{li2019spatial} &       & 93.9/144.5 & 91.6/143.1 & 88.4/140.0 &       & 58.98  & 58.85  & \textbf{30.55}~({\color{red}{$\downarrow$ 28.30}}) \\
          & SE~\cite{hu2018squeeze} &       & 89.9/140.2 & 89.9/140.2 & 79.4/127.7 &       & 49.50  & 49.00  & \textbf{21.07}~({\color{red}{$\downarrow$ 27.93}}) \\
          & DIA~\cite{huang2020dianet} &       & -     & 92.5/130.4 & 90.3/141.6 &       & -     & 51.75  & \textbf{29.43}~({\color{red}{$\downarrow$ 22.32}}) \\
    \bottomrule
    \end{tabular}%
    \end{adjustbox}
  \caption{Transfer the optimal architecture searched by EAN from image classification to crowd counting task.}\label{tab:crowdcountin}%
\end{table*}%

 \subsection{Results}
The concrete connection schemes found by EAN are presented in our supplementary. Table~\ref{tab:all_results} shows the test accuracy, the number of parameters, and relative inference time increment on CIFAR100 and ImageNet 2012. Fig.~\ref{fig:qpt} visualizes the ImageNet results from Table~\ref{tab:all_results}. Since EAN and Share-full network use sharing mechanism~\cite{huang2020dianet} for the attention module, over the vanilla ResNet, they both have fewer parameters increment than the Org-full network. Note that EAN networks have faster inference speed among the networks with the same type of attention module compared with the same type Share-full attention network. 
Furthermore, Share-full networks have higher accuracy than Org-full networks, but in most cases, the accuracy of EAN networks surpass that of Share-full networks. It implies that disconnecting the interaction between the attention and backbone in the appropriate location can maintain or even improve the performance of attention models.

\section{Analysis}
\subsection{Performance distribution of random schemes and EAN schemes}

In this part, we demonstrate that our EAN method can find effective connection schemes. For the comparison, we draw 180 random connection schemes and obtain 40 connection schemes by EAN search, both under SE module with CIFAR100. Fig.~\ref{fig:rs_vs_ean} displays the distribution of their stand-alone performance. The validation accuracy of random schemes ranges from $71$ to $75.6$ while ours EAN accuracy clusters on the right side of random scheme distribution (i.e., on the interval $[74, 76]$), which implies the EAN can readily find effective connection schemes. Specifically, EAN accuracy (average: 75.10) is greater than random schemes (average: 74.29) with P-value $4\times 10^{-7}<0.05$ under t-test. Besides, the standard derivation of EAN (std.: 0.45) is much smaller than random sampling (std.: 0.81).

\subsection{Comparison with other searching methods}
\label{sec:enas}

In this part, we compare our EAN search method with heuristic selection policy (HSP), Genetic Algorithm (GA)~\cite{Vidnerov2020Multi}, ENAS~\cite{pham2018efficient} and DARTS~\cite{liu2018darts}. HSP is a heuristic policy that makes connection every $N$ layers. For example, when $N=2$, the schemes can be $10101\cdots$ or $01010\cdots$.  Table~\ref{tab:searching} displays the experiments conducted on CIFAR100 with ResNet 164 and SE-module for different searching methods. From Table~\ref{tab:searching}, our EAN outperforms the heuristic method such as HSP or GA. Different from DARTS that searches schemes by minimizing the validation loss, the reinforcement-learning-based method~(e.g., ENAS and our method) can directly consider the validation accuracy as a reward although the accuracy or sparsity constraint is not differentiable. However, ENAS tends to converge to some periodic-alike schemes, which indicates that ENAS does not learn effective scheme from the reward.

\begin{table}[htbp]
\vspace{-0.2cm}
  \centering
  \begin{adjustbox}{width=0.5\textwidth,center}
    \begin{tabular}{cccccc}
    \toprule
    Method & Acc.  & Time Increment (\%) & Method & Acc.  & Time Increment (\%) \\
    \midrule
    EAN~(ours) & \textbf{76.93} & 23.52~({\color{red}{$\downarrow$28.57}}) & HSP~(101010..) & 75.02 & 26.62~({\color{red}{$\downarrow$25.47}}) \\
    DARTS & 75.41 & 35.03~({\color{red}{$\downarrow$17.06}}) & HSP~(010101..) & 74.87 & 26.62~({\color{red}{$\downarrow$25.47}}) \\
    GA    & 76.09 & 26.10~({\color{red}{$\downarrow$25.99}}) & HSP~(100100..) & 75.29 & 18.13~({\color{red}{$\downarrow$33.96}}) \\
     ENAS    & 76.08 & 35.07~({\color{red}{$\downarrow$17.02}}) & HSP~(010010..) & 74.01 & 18.13~({\color{red}{$\downarrow$33.96}})\\
    
    \bottomrule
    \end{tabular}%
  \end{adjustbox}
  \caption{Comparison of the testing accuracy and relative inference time increment of the searched network for different methods.}
  \label{tab:searching}%
\end{table}%

From our empirical results, the controller of ENAS tends to converge to some periodic-alike schemes at a fast speed. In this case, it will conduct much less exploration of the potential efficient structures. The majority of the schemes searched by ENAS are ``111...111''~(Share-full network) or ``000...000''~(Vanilla network), which shows that it can not get the balance between the performance and inference time. The list of schemes searched by ENAS is presented in our supplementary. In Table~\ref{tab:ean_vs_enas}, the minority of the periodic-alike schemes searched by ENAS are shown, e.g., ``001'' in ENAS~(a). Such schemes may come from the input mode of ENAS, 
\textit{i.e.}, for a connection scheme $\mathbf{a} = (a_1,a_2,...,a_m)$, the value of component $a_l$ depends on $a_{l-1},a_{l-2},...,a_1$. Such strong sequential correlations let the sequential information dominate in the RNN controller instead of the policy rewards. Compared with the periodic-alike connection schemes from ENAS, the schemes from EAN demonstrate better performance.

Besides, our experiment indicates that ENAS explores a much smaller number of candidate schemes. We quantify the convergence of the controller using $\bar{\mathbf{p}}=\frac{1}{m}\sum_{i=1}^m\hat{p}_\theta^i$, which is the mean of the probability $\hat{\mathbf{p}}$ associated with the scheme. When $\bar{\mathbf{p}}$ is close to 1, the controller tends to generate a deterministic scheme. Fig.~\ref{fig:enas_vs_ean} shows the curve of $\bar{\mathbf{p}}$ with the growth of searching iterations, where $\bar{\mathbf{p}}$ of ENAS shows the significant tendency for convergence in 20 iterations and converges very fast within 100 iterations. Generally speaking, methods in NAS~\cite{pham2018efficient,zoph2016neural} require hundreds or thousands of iterations for convergence.

\subsection{Transferring Connection Schemes}
To further investigate the generalization of EAN, we conduct experiments on transferring the optimal architecture from image classification to crowd counting task~\cite{zhang2016single,cao2018scale,li2018csrnet,hossain2019crowd}, semantic segmentation~\cite{Everingham10}.

\textbf{Crowd counting.} Crowd counting aims to estimate the density map and predict the total number of people for a given image, whose efficiency is also crucial for many real-world applications, \textit{e.g.}, video surveillance and crowd analysis. However, most state-of-the-art works still rely on the heavy pre-trained backbone networks~\cite{liu2020efficient} for obtaining satisfactory performance on such dense regression problems. The experiments show that the network obtained by EAN on ImageNet serves as an efficient backbone network and can extract the representative features for crowd counting. The networks pre-trained on the ImageNet dataset serve as the backbone of crowd counting models. We evaluate the transferring performance on the commonly-used Shanghai Tech dataset~\cite{zhang2016single}, which includes two parts. Shanghai Tech part A~(SHHA) has 482 images with 241,677 people counting, and Shanghai Tech part B~(SHHB) contains 716 images with 88,488 people counting. Following the previous works, SHHA and SHHB are split into train/validation/test set with 270/30/182 and 360/40/316 images, respectively. The performance on the test set is reported using the standard Mean Square Error~(MSE) and Mean Absolute Error~(MAE), as shown in Table~\ref{tab:crowdcountin}. 
Our EAN can outperform the baseline (Org-full and Share-full) while reducing the inference time increment by over 40\% compared with the baseline.

\begin{figure}[t]
    \centering
    \includegraphics[width=\linewidth]{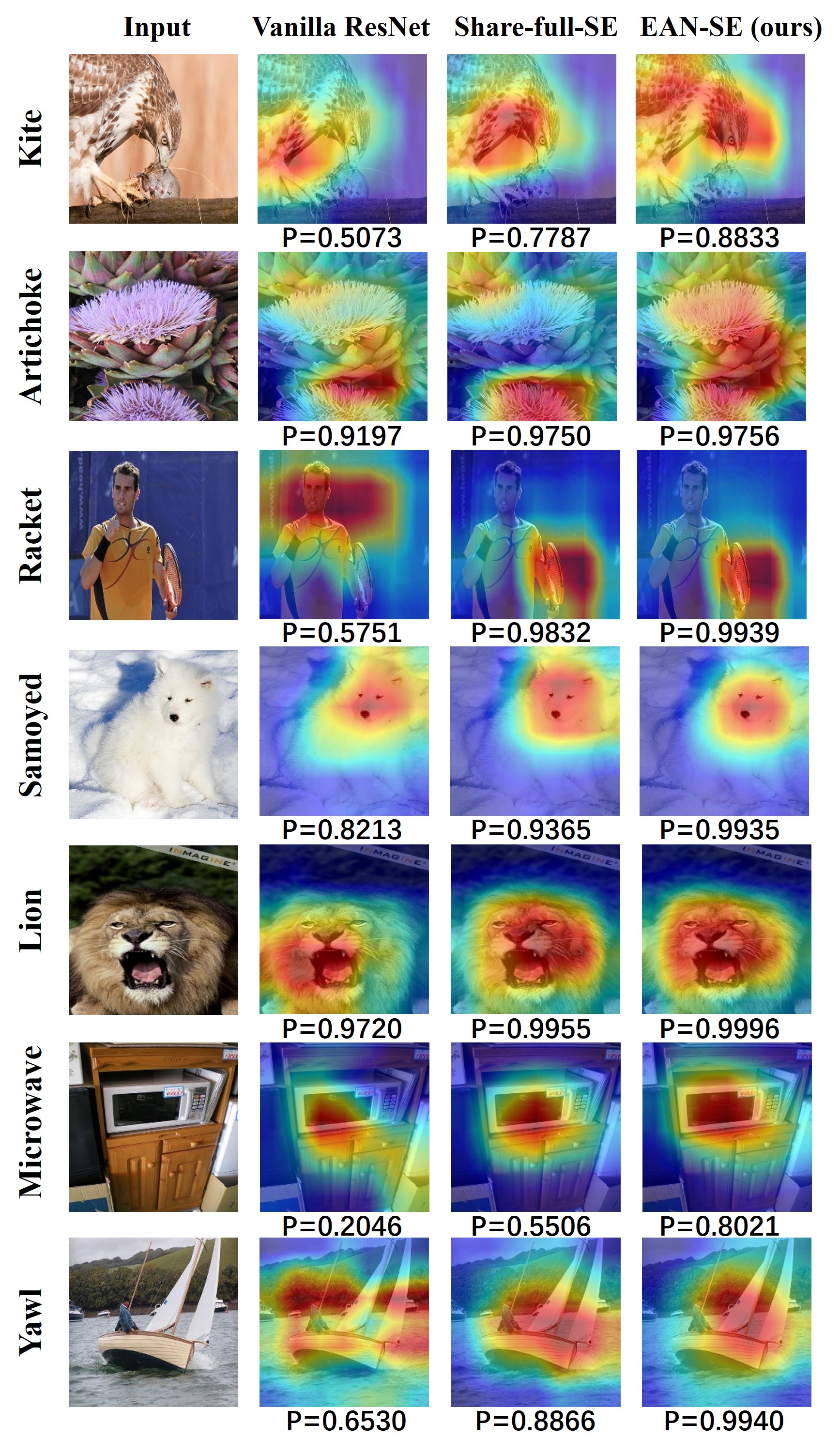}
    \caption{Grad-CAM visualization of different attention models. The red region indicates an essential place for a network to obtain a target score ($\mathbf{P}$) while the blue region is the opposite.}
    \label{fig:cam}
\end{figure}

\textbf{Semantic segmentation.}
We verify the transferability of the attention networks obtained by EAN on semantic segmentation task in Pascal VOC 2012~\cite{everingham2015pascal} dataset. Table~\ref{tab:semseg} shows the performance comparison of the backbone with different attention modules, e.g., DIA and SE. Again, our results indicate that the EAN-network can maintain the performance of the Share-full network and significantly reduce the time increment compared with the Share-full network, which shows EAN has the capacity of transferring to semantic segmentation.
\begin{table}[htbp]
  \centering
    \begin{adjustbox}{width=0.45\textwidth,center}
    \begin{tabular}{ccc}
    \toprule
    Model & mIoU/mAcc/allAcc~(\%) & Time Increment~(\%)\\
    \midrule
    Org-ResNet & 69.39 / 78.87 / 92.97&- \\
    Share-full-SE & 73.03 / 82.13 / 93.74&48.16 \\
    EAN-SE & 73.68 / 83.08 / 93.79& \textbf{16.43}~({\color{red}{$\downarrow$ 31.73}}) \\
    Share-full-DIA & 74.02 / 83.11 / 93.92&64.86 \\
    EAN-DIA & 73.91 / 82.93 / 93.92 & \textbf{7.68}~({\color{red}{$\downarrow$ 57.18}})\\
    \bottomrule
    \end{tabular}%
    \end{adjustbox}
  \caption{Performance and relative inference time increment comparison on Pascal VOC 2012 val set.}
  \label{tab:semseg}%
  \vspace{-0.3cm}
\end{table}%

\subsection{Capturing Discriminative Features}
To study the ability of EAN in capturing and exploiting features of a given target, we apply Grad-CAM~\cite{selvaraju2017grad} to compare the regions where different models localize on with respect to their target prediction. Grad-CAM is a technique to generate the heatmap highlighting network attention by the gradient related to the given target. Fig.~\ref{fig:cam} shows the visualization results and the softmax scores for the target with vanilla ResNet50, Share-full-SE, and EAN-SE on the validation set of ImageNet 2012. The red region indicates an essential place for a network to obtain a target score while the blue region is the opposite. The results show that EAN-SE can extract similar features as Share-full-SE, and in some cases, EAN can even capture much more details of the target associating with higher confidence for its prediction. This implies that the searched attention connection scheme may have a more vital ability to emphasize the more discriminative features for each class than the two baselines ~(Vanilla ResNet and Share-full-SE). Therefore it is reasonable to bring additional improvement on the final classification performance with EAN in that the discrimination is crucial for the classification task, which is also validated from ImageNet test results in Table~\ref{tab:all_results}.

\section{Conclusion}
To improve the efficiency of using the attention module in a network, we propose an effective EAN method to search for an optimal connection scheme to plug the modules. Our numerical results show that the attention network searched by our method can preserve the original accuracy while reducing the extra parameters and accelerating the inference. We empirically illustrate that our attention networks have the capacity of transferring to other tasks and capturing the informative features.

\end{document}